\documentclass[letterpaper]{article} 
\usepackage{aaai24}  
\usepackage{times}  
\usepackage{helvet}  
\usepackage{courier}  
\usepackage[hyphens]{url}  
\usepackage{graphicx} 
\usepackage{natbib}  
\usepackage{caption} 
\usepackage{algorithm}

\usepackage{newfloat}
\usepackage{listings}
\DeclareCaptionStyle{ruled}{labelfont=normalfont,labelsep=colon,strut=off} 
\floatstyle{ruled}
\newfloat{listing}{tb}{lst}{}
\floatname{listing}{Listing}
\title{Quantile-Based Maximum Likelihood Training for Outlier Detection}
\author{
    Masoud Taghikhah\textsuperscript{\rm 1}\equalcontrib,
    Nishant Kumar\textsuperscript{\rm 1}\equalcontrib\thanks{Corresponding author.},
    Sini\v{s}a \v{S}egvi\'{c}\textsuperscript{\rm 2},
    Abouzar Eslami\textsuperscript{\rm 3},
    Stefan Gumhold\textsuperscript{\rm 1}
}
\affiliations{
    \textsuperscript{\rm 1}Faculty of Computer Science, Technische Universität Dresden, Dresden, Germany
\\

    \textsuperscript{\rm 2}Faculty of Electrical Engineering and Computing, University of Zagreb, Zagreb, Croatia
    \\
    \textsuperscript{\rm 3}Translational Research Lab, Carl Zeiss Meditec AG, Munich, Germany
    \\
    
    {masoud.taghikhah, nishant.kumar, stefan.gumhold}@tu-dresden.de, sinisa.segvic@fer.hr, abouzar.eslami@zeiss.com
}

\usepackage{bibentry}

\begin{document}

\maketitle

\begin{abstract}
Discriminative learning effectively predicts true object class for image classification.~However, it often results in false positives for outliers, posing critical concerns in applications like autonomous driving and video surveillance systems.~Previous attempts to address this challenge involved training image classifiers through contrastive learning using actual outlier data or synthesizing outliers for self-supervised learning.~Furthermore, unsupervised generative modeling of inliers in pixel space has shown limited success for outlier detection.~In this work, we introduce a quantile-based maximum likelihood objective for learning the inlier distribution to improve the outlier separation during inference.~Our approach fits a normalizing flow 
to pre-trained discriminative features and detects the outliers according to the evaluated log-likelihood.~The experimental evaluation demonstrates the effectiveness of our method as it surpasses the performance of the state-of-the-art unsupervised methods for outlier detection.~The results are also competitive compared with a recent self-supervised approach for outlier detection.~Our work allows to reduce dependency on well-sampled negative training data, which is especially important for domains like medical diagnostics or remote sensing.

\end{abstract}

\section{Introduction}

As AI-based fully-autonomous technologies get introduced in several commercial sectors, their reliability becomes important in safety-critical systems such as self-driving cars and surgical robotics.~For instance, popular image classifiers~\cite{Xie2016,7780459} detect inlier objects with high accuracy.~However, they suffer from degraded performance while detecting outlier objects~\cite{hendrycks2016baseline}. 
~Hence, it is vital to train models that can precisely learn the data distribution of inliers such that anomalous instances can be detected as an outlier.~Such a model should significantly improve the reliability of fully-autonomous systems for safety-critical applications while satisfying evolving regulatory requirements and reinforcing the public's trust in such systems for real-world deployment~\cite{avs}.


There are a lot of outlier-aware image classification approaches that exist in the literature.~They can be categorized into supervised~\cite{DBLP:conf/iclr/HendrycksMD19}, unsupervised~\cite{hendrycks2016baseline, lee2018simple, liang2018enhancing, liu2020energy}, and self-supervised~\cite{tack2020csi, du2022vos}.~A recent approach~\cite{du2022vos} generates synthetic features from the learned inlier feature distribution and selects a subset of those features as outliers based on a pre-defined sampling strategy.~Afterward, the selected outlier features are used for energy-based regularization of the final layer of the classification network to encourage a lower energy for inlier features and higher for outlier features.~Subsequently, during model inference, \cite{du2022vos} perform outlier detection by thresholding the energy scores computed from the classification head to discriminate outlier instances from inliers.

\begin{figure*}[!htb]
  \centering
\includegraphics[width=0.98\linewidth]{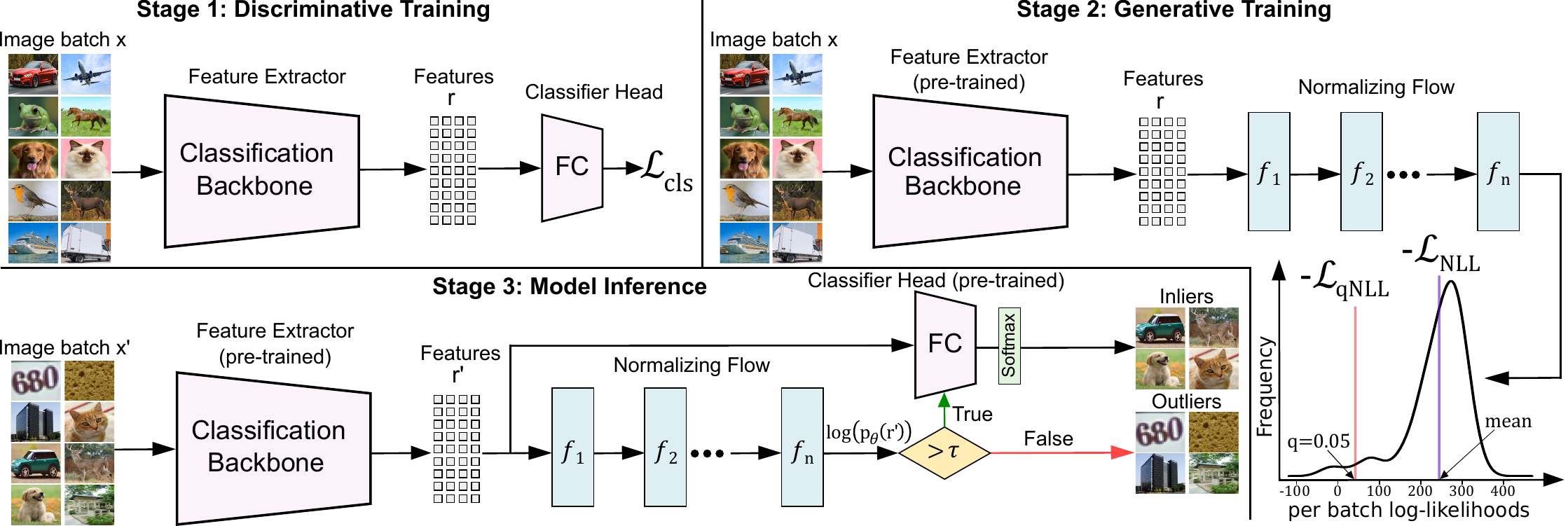}
\caption{Overview of the \texttt{QuantOD} framework during two-stage training and single-stage inference.~During discriminative training, the end-to-end image classification model including the fully-connected (FC) head is trained with standard cross-entropy loss~$\mathcal{L}_{cls}$.~In the generative training stage, the inlier features $r$ are extracted from the pre-trained classification model and used to train the normalizing flow model with the negative log-likelihood loss $\mathcal{L}_{qNLL}$.
~Note that the parameters of the classification backbone network are frozen during generative training.~During inference, a test image $x'$ is labeled as an outlier if its likelihood score obtained from the flow model is less than a pre-defined likelihood threshold $\tau$.}
   \label{fig:framework}
\end{figure*}

Maximum likelihood training of the inlier data distribution should assign a lower probability estimate to the outliers.~Generative models like normalizing flows offer precise density estimation capability for representing complex  distributions~\cite{kingma2018glow}.~However, prior works~\cite{NFlows_fails, nalisnick2018deep} showed that in the pixel space, the density learned by generative models could not distinguish anomalous instances from inliers as such learning scheme seems to fail to capture the semantic contents.~To overcome this problem, we propose a novel framework named $\texttt{QuantOD}$~(Quantile-based Maximum Likelihood Training for Outlier Detection) that derives the inlier feature representations from a pre-trained image classifier and estimates
its density using a new maximum likelihood objective that
leverages the $q$-quantile of the log-likelihood scores per
training batch for learning the generative model.~Our framework determines the log-likelihood-based threshold by analyzing the likelihoods of inliers from the validation set.~Subsequently, $\texttt{QuantOD}$ performs outlier detection by examining whether the log-likelihood of a test feature is above or below this threshold.~Unlike methods like~\cite{du2022vos} that synthesize outlier features for classifier regularization, $\texttt{QuantOD}$ trains exclusively on inlier data.~The conceptual simplicity of our method decreases the likelihood of undesirable effects 
such as bias and over-optimistic evaluation.

Quantile-based optimization has been previously studied in Reinforcement Learning~\cite{dabney2018distributional, zhou2020non, 10.5555/3586210.3586438} and for Hyperparameter tuning~\cite{pmlr-v119-salinas20a, salinas-icml23}.~However, the optimization of a generative model based on the likelihood estimate of the inliers near the boundary of the distribution is largely unexplored.~During training, \texttt{QuantOD} passes the input image to the backbone network and extracts the features from an appropriate network layer.~Our framework can adapt to backbone networks such as ResNet~\cite{7780459} or Vision Transformers~\cite{dosovitskiy2021an} for image classification.~Subsequently, the features from the backbone network are provided to our generative model to estimate the density of inlier feature representations.~The approach involves learning a generative model based on normalizing flow with a maximum likelihood objective for inlier features.~During inference, $\texttt{QuantOD}$ detects the outliers by comparing their likelihoods to the threshold we validate on inliers. The main contributions of our work are as follows:

\begin{itemize}
\item 
We study a principled approach for outlier-aware image classification based on the log-likelihood scoring of the pre-trained discriminative features.

\item 
We propose to increase the separation between inliers and outliers by introducing a robust quantile-based maximum likelihood objective for learning a generative model of the inlier distribution.

\item Our method outperforms the state-of-the-art for unsupervised outlier detection on publicly available datasets while preserving the inlier classification performance.

\item We report similar inference runtime of our approach compared with other unsupervised outlier detection methods due to the minimal computational overhead of our generative model. 
\end{itemize}

\section{Related Work}
Recent methods like Outlier Exposure~\cite{DBLP:conf/iclr/HendrycksMD19}, ATOM~\cite{10.1007/978-3-030-86523-8_26} and Natural Habitats~\cite{pmlr-v162-katz-samuels22a}, employ publicly available outlier datasets in a supervised training setup.~However, defining outliers, in general, is challenging, as any data point not sampled from the inlier data distribution can be regarded as an outlier.~Hence, utilizing outlier datasets for model supervision might lead to sub-optimal results for unknown outliers.~Therefore, the following study exclusively focuses on self-supervised and unsupervised methods for outlier detection.

\paragraph{Self-supervised Outlier Detection:} 
Prior work like~\cite{lee2018training} used GANs to generate outliers in pixel space and performed contrastive training, while \cite{tack2020csi} perturbed inlier instances to distinguish it from the non-perturbed instances.~Similarly, ALOE~\cite{chen2022robust} trained the model with adversarial outliers generated within the $\epsilon$-ball.~Other methods~\cite{mohseni2020self,hendrycks2020augmix,DBLP:conf/nips/HendrycksMKS19, 9857016, pmlr-v162-ming22a, tao2023nonparametric, du2022vos} proposed to regularize image classifiers via self-supervised training.~However, synthetic outliers may be sampled from imprecise decision boundaries.~Works such as~\cite{Rudolph_2021_WACV,Rudolph_2022_WACV,Rudolph_2023_WACV} performed defect detection of industrial images by estimating the density of inlier features using normalizing flow.~\cite{Li_2021_CVPR} cut an image patch from one location and randomly pasted it so that the model detects features of perturbed patches.~\cite{Ding_2022_CVPR} learn features of reference inlier images and their pseudo anomalies to obtain an anomaly score, while \cite{Deng_2022_CVPR} presents an encoder-decoder model where the one class embedding of the encoder preserves information about standard patterns but forgets anomalous perturbations.~\cite{spot_difference} increases the model's invariance to global changes in the input image by forcing dissimilarity between the image features and its local perturbations, whereas~\cite{Sheynin_2021_ICCV} captures the multi-scale patch distribution of each training image and performs image transformations to distinguish between natural and transformed image patches.~In contrast,~\cite{sparse_representation} exploit self-supervised sparse representation of inlier features for anomaly detection.~Note that defect detection is a distinct task compared to image classification; the former locates defects in an image, whereas the latter labels the entire image as an outlier or inlier.

\paragraph{Unsupervised Outlier Detection:}Works like~\cite{hendrycks2016baseline} use softmax function to detect outliers while~\cite{liang2018enhancing,hsu2020generalized} employ image perturbations to enhance the performance of softmax function.~Others use Mahalanobis distance~\cite{lee2018simple}, \textit{k}-nearest neighbor~\cite{pmlr-v162-sun22d}, Rectified Activations~\cite{sun2021react}, KL divergence~\cite{huang2021importance}, DICE~\cite{10.1007/978-3-031-20053-3_40}, Gram Matrices~\cite{DBLP:conf/icml/SastryO20} and Energy~\cite{liu2020energy, lin2021mood, wang2021canmulti} instead of softmax function.
~Some methods perform outlier detection by leveraging both inlier features and image reconstruction error~\cite{zong2018deep}, estimating the Gaussian mixture model of inlier features~\cite{morteza2022provable}, or performing Kolmogorov-Smirnov test on the flow's latent representations~\cite{jiang2022revisiting}.~\cite{9412109} represent features as multivariate Gaussians for anomaly detection while~\cite{zhang00eccv, DBLP:journals/ijcv/BlumSNSC21} use embedded features to train the flow model for pixel-level and image-level outlier detection.~\cite{Gudovskiy_2022_WACV} use conditional flows to learn inlier features for defect detection in industrial images while~\cite{Roth_2022_CVPR} and \cite{9839549} use inlier feature embeddings from the pre-trained model as a memory bank.~PaDiM~\cite{padim} estimates inlier patch embeddings as multivariate Gaussians and localize anomalies using Mahalanobis distance while~\cite{Tsai_2022_WACV} learns multi-scale patches to differentiate the anomalous regions.
~However, with significant class imbalance, accurate outlier detection becomes challenging for unsupervised methods.~Also, they are highly sensitive to hyperparameters, necessitating precise tuning, which is a demanding task without a labeled validation set.~Importantly, previous self-supervised or unsupervised methods have not explored the quantile function for robust maximum likelihood training of the density estimator for outlier-aware image classification.

\section{Method}
The key objective of an open-set image classifier is to predict the correct class of an inlier image accurately
while being able to reject the outlier images.~We denote the $N$ training images as $x \sim \mathcal{X}$, and the ground truth class labels for each inlier image as $y \sim \mathcal{Y}$ where each label refers to one of the $K$ known classes.~During inference, we extract features of an image $x$ and estimate the density using our flow-based generative model.~In the subsequent discussion, we provide an overview of our approach and present the quantile-based maximum likelihood objective for density estimation of the inlier feature representations.~We also describe the model inference and thresholding scheme for outlier detection.

\subsection{QuantOD: An Overview}
\label{overview}
The workflow of our framework is shown in Figure~\ref{fig:framework}.~In the first stage, we train a standard image classifier using the multi-class cross-entropy loss $\mathcal{L}_{cls}$ given the ground-truth class labels $y$.~In the second stage, we obtain the inlier features $g(x)$ (denoted as $r$) from the penultimate layer of the pre-trained image classifier.~Note that $g$ is a learnable function which we term as the backbone of the classification network that transforms inlier images $x$ to fixed-size inlier features $g(x)$.~We then estimate the probability distribution of $g(x)$ via generative modeling using the invertible normalizing flow network.~The flow network transforms the unknown distribution of inlier features $g(x)$ into a standard normal distribution through the negative log-likelihood loss $\mathcal{L}_{qNLL}$ where $0 \leq q < 1$ is the desired quantile.~Finally, given a test image during the model inference, we extract the features from the classifier and compute the log-likelihood score of the test image using the pre-trained flow network.~The decision whether the test image is an inlier or an outlier depends on the log-likelihood-based threshold $\tau$. If the log-likelihood score of the test feature is greater than $\tau$, the test feature is passed through the classification head to estimate the inlier object class along with the softmax confidence score.~If the log-likelihood score of the test feature is below $\tau$, we assign the test image as an outlier.

\subsection{Density Estimation of Inlier Features}
\label{main_method}
The task of the density estimator in the $\texttt{QuantOD}$ framework is to model the probability distribution $p(g(x))$ of the inlier features.~A robust density estimator should recognize whether a test feature is likely or unlikely to be an outlier.~We specify our probabilistic density estimator as a parametric function $f$ to facilitate the selection of common distributions such as standard Gaussians and subsequently estimate the parameters of $f$ using the inlier features $g(x)$.~Suppose $g(x)$ follows a multivariate normal distribution, meaning $g(x)$ is modeled as $\mathcal{N}(\mu, \Sigma)$ where $\mu$ is the mean and $\Sigma$ is the covariance of $g(x)$.~Then, such a density estimator $f$ can compute the maximum likelihood estimates based on the parameters $\mu$ and $\Sigma$.~On the other hand, if the distribution of $g(x)$ is assumed to be unknown, then $f$ can be learned as a trainable neural network, such as flows that transforms this unknown distribution into multivariate Gaussian and allows efficient computation of log-likelihoods. Hereon, we will use the term $g(x)$ as $r$ to simplify our problem formulation.  

\paragraph{Normalizing Flow-based Distribution Learning:}
\label{nf_based}
Let us define the probability distribution $p(r)$ of inlier features as unknown with $r \in \mathcal{R}$. We use normalizing flows as a learnable density estimator $f$ that estimates $p(r)$ indexed with learnable parameters $\theta \in \Theta$.~We formulate the latent observations of the flow as $z \in \mathcal{Z}$ with probability distribution $p(z)$ defined as multivariate standard normal distribution such that \(p({z}) = \mathcal{N}(0, \,I)\) with zero mean and unit covariance.~Due to the excellent property of the normalizing flow network to be an invertible and bijective mapping between $\mathcal{R}$ and $\mathcal{Z}$, it transforms $p(r)$ into a normally distributed $p(z)$ in the latent space.~Hence, the transformation \(f:\mathcal{R} \rightarrow \mathcal{Z}\) is deterministic and preserves the dimensionality of the inlier features in the latent space such that $f:{R}^m \to {R}^m$ where $m$ are the feature dimensions.

\paragraph{Network Architecture:}
 The basic element of our normalizing flow network $f$ is a series of $n$ invertible and bijective mappings called coupling blocks such that $f:= (f_1,..,f_j,...,f_n)$ with $r = f^{-1}(z; \theta)$ and $z = f(r;\theta)$.~We follow~\cite{dinh2017density} to construct our coupling blocks as learnable affine transformations, i.e., scaling $s$ and translation $t$, respectively.~We associate $s$ and $t$ to be fully connected neural networks.
 ~The input to an arbitrary $j^{th}$ coupling block is first split into two parts $u^{j}_1$ and $u^{j}_2$ that are transformed by learnable $s_1, t_1$ and $s_2, t_2$ networks and are coupled alternatively.~The output of the $j^{th}$ block is the concatenation of the resulting parts $v^{j}_1$ and $v^{j}_2$ given as:

 \begin{equation}
     v^{j}_1 = u^{j}_1 \odot \exp \big(s_2(u^{j}_2)\big) + t_2(u^{j}_2)   
    \label{eq1}
 \end{equation}
 
  \begin{equation}
    v^{j}_2 = u^{j}_2 \odot \exp \big(s_1(v^{j}_1)\big) + t_1(v^{j}_1) 
    \label{eq2}
 \end{equation}
 
where $\odot$ is the element-wise multiplication, and the exponential function ensures non-zero coefficients.~The benefit of constructing such a transformation is that we can easily recover $u^{j}_1, u^{j}_2$ from $v^{j}_1,v^{j}_2$ in the inverse direction of this coupling block with subtle modifications in the architecture. Note that $s$ and $t$ need not be invertible and can be represented by any arbitrary neural network. Additionally, we follow~\cite{kingma2018glow} to perform a learned invertible $1 \times 1$ convolution operation after every coupling block to reverse the ordering of the feature channels, thereby ensuring each feature influences each other through the transformation. 

\subsection{Quantile-based Maximum Likelihood Training}
We aim to learn the parameters $\theta$ of the flow network such that the unknown probability distribution of the inlier features $p(r)$ is transformed into a standard normal distribution $p(z)$.~According to the change of variables formula, we can define the posterior distribution \(p_{\theta}({r})\) as,

\begin{equation}
p_{\theta}({r}) =  p(f(r; \theta)) *
  \left|
    \mathrm{det} \frac{
      \partial f(r; \theta)
    }{
      \partial {r}^T\
    }
  \right| \label{eq3} \end{equation}


As the training loss requires a minimization objective, we can train the network using the negative average of the log-likelihood scores for a set of samples in each training batch.~However, the arithmetic mean provides a measure of central tendency in a batch of log-likelihood scores such that some inliers may receive very low likelihoods.~Therefore, we propose a negative log-likelihood loss, $\mathcal{L}_{qNLL}$, that is the negation of the $q$-quantile of the log-likelihood scores of inlier features for each training batch.~Given the observed samples in a training batch, the flow parameters $\theta$ are optimized by the negative log-likelihood of $p_\theta(r)$ as: 


\begin{equation}
\mathcal{L}_{qNLL}(r;\theta) = - \mathcal{Q}_{q}\left(\log(p_\theta(r_{ij}))\right)_{i,j=1}^{N}
\label{eq4} 
\end{equation}

where \(r_{ij}\) represent the inlier features of the $i^{th}$ training image sample in the $j^{th}$ training batch, and $\mathcal{Q}_{q}$ represents the quantile function that first sorts the log-likelihood scores in the training batch in ascending order and subsequently selects the log-likelihood score at $q$-quantile.~If the $q$-quantile lies between the log-likelihood scores of two samples, the result is computed by linearly interpolating between the two values.~If we select a low $q$ (e.g., $q = 0.05$), the loss formulation in Eq.~\ref{eq4} will encourage the likelihood maximization of the low-likelihood inlier samples located at the distribution boundary.~Therefore, after training flow network $f$ and updating the parameters $\theta$ until the convergence of the $\mathcal{L}_{qNLL}$ loss, the likelihood $p_\theta(r)$ of inlier features is likely to be higher than those of outlier features.

\subsection{Log-Likelihood Based Threshold}
\label{inference}
During inference, the log-likelihood score of a test feature $r'$ can be derived from the learned flow model with parameters $\theta$.~The feature $r'$ to be assigned as an inlier or an outlier relies on the log-likelihood score $\log(p_{\theta}(r'))$.~Since it is a binary classification problem, a threshold $\tau$ is required to distinguish inlier and outlier features.~We assign test feature $r'$ as an outlier if $\log(p_{\theta}(r')) < \tau$ and inlier if $\log(p_{\theta}(r')) \geq \tau$.~If the test feature is detected as an inlier, we obtain the predicted inlier class with the softmax confidence score from the classification head.~We label the test feature as an outlier if  $\log(p_{\theta}(r')) < \tau$.~For fixing a suitable threshold $\tau$, there is a natural trade-off between false positives (i.e., outliers wrongly classified as inliers) and false negatives (i.e., inliers misclassified as outliers).~Therefore, we adopt the TPR-$\beta$ thresholding scheme for determining $\tau$, where $1 - \beta$ is a false alarm rate and $\beta$ is fixed at 95\%, meaning 95\% of the inlier features are correctly detected.~We evaluate the performance of our approach using the standard outlier detection metrics such as Area under the Receiver Operating Characteristics (AUROC  $\uparrow$), False Positive Rate at 95\% True Positive Rate (FPR95 $\downarrow$), and Area under Precision-Recall Curve (AUPR  $\uparrow$) where True Positive is the correct detection of an inlier image.~For inlier classification performance, we compute accuracy between the predicted and groundtruth classes for the validation images of the inlier dataset.

\begin{table*}[!ht]
    \centering
\scalebox{0.92} {\begin{tabular}{|c|ccccccc|c|}
    \hline
         \textbf{Method} & \multicolumn{7}{c|}{\textbf{Outlier Datasets (Metrics: FPR95  $\downarrow$ (\%) / AUROC $\uparrow$ (\%))}} \\ \hline
        & Textures & SVHN & Places365 & LSUN-C & LSUN-R & iSUN & Average  \\ 
           \cline{2-8}
       MSP~\shortcite{hendrycks2016baseline} & 59.28 / 88.50 & 48.49 / 91.89 & 59.48 / 88.20 & 30.80 / 95.65 &  52.15 / 91.37 & 56.03 / 89.83 & 51.04 / 90.91  \\ 
       Mahal~\shortcite{lee2018simple} & 15.00 / \textbf{97.33} & 12.89 / 97.62 & 68.57 / 84.61 & 39.22 / 94.15 &  42.62 / 93.23 & 44.18 / 92.66 & 37.08 / 93.27  \\ 
       ODIN~\shortcite{liang2018enhancing} & 49.12 / 84.97  & 33.55 / 91.96 & 57.40 / 84.49 & 15.52 / 97.04 &  26.62 / 94.57 & 32.05 / 93.50 &  35.71 / 91.09  \\ 
        Energy~\shortcite{liu2020energy} & 52.79 / 85.22  & 35.59 / 90.96 & 40.14 / 89.89 & 8.26 / 98.35 &  27.58 / 94.24 & 33.68 / 92.62 & 33.01 / 91.88  \\ 
       GEM~\shortcite{morteza2022provable} & 15.06 / 97.33  & 13.42 / 97.59  & 68.03 / 84.44  & 39.46 / 94.13  &  42.89 / 93.27 & 44.41 / 92.60 & 37.21 / 93.23 \\
        \textbf{QuantOD (Ours)} & \textbf{12.85} / 97.27 & \textbf{9.39} / \textbf{97.94} & \textbf{36.75} / \textbf{91.04} & \textbf{7.49} / \textbf{98.42} & \textbf{19.05} / \textbf{96.08} & \textbf{17.50} / \textbf{96.39} & \textbf{17.17} / \textbf{96.19}  \\
        
    \hline
    
         \textbf{Method} & \multicolumn{7}{c|}{\textbf{Metrics: AUPR  $\uparrow$ (\%) / Inference Time $\downarrow$ (seconds)}}  \\ \hline
        & Textures & SVHN & Places365 & LSUN-C & LSUN-R & iSUN & Average \\ 
           \cline{2-8}
       MSP~\shortcite{hendrycks2016baseline} & 97.16 / \textbf{15.15} & 98.27 / 6.22 & 97.10 / 17.23 & 99.13 / 7.96 & 98.12 / 13.79 & 97.74 / 9.05 & 97.92 / 11.57  \\ 
       Mahal~\shortcite{lee2018simple} & \textbf{99.41} / 61.03 & 99.47 / 51.02 & 96.20 / 54.95 & 98.81 / 52.76 &  98.60 / 74.99 & 98.45 / 58.69 & 98.49 / 58.91   \\ 
       ODIN~\shortcite{liang2018enhancing} & 95.28 / 19.89  & 98.00 / 10.64 & 95.82 / 19.76 & 99.33 / 11.83 &  98.77 / 15.10 & 98.54 / 12.83 & 97.62 / 15.01   \\ 
        Energy~\shortcite{liu2020energy} & 95.41 / 15.49  & 97.64 / \textbf{6.08} & 97.30 / 15.53 & 99.66 / \textbf{7.82} &  98.67 / 9.26 & 98.27 / \textbf{7.90} & 97.83 / \textbf{10.35}  \\ 
       GEM~\shortcite{morteza2022provable} & 99.41 / 61.03   & 99.47 / 51.02  & 96.11 / 54.95  & 98.81 / 52.76   &  98.61 / 74.99  & 98.42 / 58.69  & 98.47 / 58.91 \\
        \textbf{QuantOD (Ours)} & 99.40 / 18.88   & \textbf{99.59} / 7.53  & \textbf{97.73} / \textbf{11.28} & \textbf{99.70} / 10.35  & \textbf{99.20} / \textbf{9.16}  & \textbf{99.28} / 9.09  & \textbf{99.15} / 11.04   \\
        \hline
    \end{tabular}
    }
    \caption[]{Validation performance of \texttt{QuantOD} framework compared with five recent unsupervised methods for outlier detection given the image classification task on CIFAR-10 as the inlier dataset.~The performance for other methods are courtesy of Energy~\cite{liu2020energy} and GEM~\cite{morteza2022provable}.~Each of the methods were trained on WideResNet40~\cite{DBLP:conf/bmvc/ZagoruykoK16} as the backbone architecture.~The results from \texttt{QuantOD} are averaged from three trial runs each with different initial random seeds.~The inlier classification accuracy for other methods as reported by Energy~\cite{liu2020energy} is 94.84\%, and we report 94.68\% for the \texttt{QuantOD} framework.~We report the inference runtime for all the compared methods on the same hardware settings while running the pre-trained model on the outlier datasets. } 
    \label{tab:sota}
\end{table*}

\section{Experiments}
This section provides experimental details about the proposed $\texttt{QuantOD}$ framework and outlines the image datasets used during training and model inference.~We also show our main results on the outlier detection performance and compare them with other related methods on different combinations of inlier and outlier datasets.~Subsequent experiments involve ablation studies, where we analyze the effect of the meta-parameters, and the network architecture of the image classifier as well as the normalizing flow network on outlier detection.~We also report the results for the inlier classification performance and inference runtime. 

\paragraph{Datasets:}\label{data} We train $\texttt{QuantOD}$ on  CIFAR-10~\cite{krizhevsky2009learning} dataset that consists of 60000 color images of $32 \times 32$ resolution in 10 different object classes, with 6000 images per class.~The dataset split has 50000 training images and 10000 test images and are evenly distributed among the classes, each representing an object.~We evaluate the performance of our model on six outlier datasets, namely Textures~\cite{6909856}, SVHN~\cite{netzer2011reading}, Places365~\cite{zhou2017places}, LSUN-C~\cite{DBLP:journals/corr/YuZSSX15}, LSUN-R~\cite{DBLP:journals/corr/YuZSSX15} and iSUN~\cite{DBLP:journals/corr/XuEZFKX15}.~The outlier datasets were selected such that there was no overlap between inlier object classes and outlier semantics. 

\paragraph{Classification network:}
\label{model} 
For training our $\texttt{QuantOD}$ framework with the inlier CIFAR-10 dataset, we used WideResNet40~\cite{DBLP:conf/bmvc/ZagoruykoK16} as the backbone architecture for the image classification with a depth and width of $40$ and $2$, respectively.~The architecture is also the backbone network for the other evaluated approaches in Table~\ref{tab:sota}.~In our work, the meta-parameters of the WideResNet40 model were fixed similar to the ones used by Energy~\cite{liu2020energy}.~The features used for training our normalizing flow model are 128-dimensional that are obtained from the penultimate layer of the WideResNet40 network.~During inference, the extracted features are utilized by the density estimator for likelihood-based outlier detection and are also used by the final fully-connected layer to predict the inlier classification logits.  

\paragraph{Density estimator:}
\label{density}We use normalizing flow as the density estimator to model the distribution of inlier features.~The results in Table 1 are based on the  Glow~\cite{kingma2018glow} architecture that performs a learned invertible $1 \times 1$ convolution after every coupling block such that each feature influences the learnable parameters of the flow model.~The $s$ and $t$ transformations are fully-connected (FC) networks, and we do not perform sub-sampling on the features, thereby preserving their dimensionality.

\paragraph{Meta-parameters:}
The training and inference of the $\texttt{QuantOD}$ framework were done using four NVIDIA A100-SXM4 GPUs.~For training our framework, the standard training split of the inlier dataset is utilized in the experimental setup.~In the first stage, the classifier was trained for 100 epochs, with standard Cross-Entropy loss using Stochastic Gradient Descent (SGD) optimization with a learning rate of $0.1$, the momentum of $0.9$, and a weight decay of $5\mathrm{e}{-4}$.~We also use the cosine learning rate scheduler~\cite{loshchilov2017sgdr}, which gradually decays the initial learning rate of $0.1$ to a minimum of $1\mathrm{e}{-6}$.~The dropout rate of the classifier was set to $0.3$.~The second stage flow model was trained for $50$ epochs with the dropout rate for the fully-connected network in each coupling block set as 0.3 and Adam optimizer having a learning rate of $9\mathrm{e}{-5}$ and weight decay of $1\mathrm{e}{-6}$.~The batch size for both the first and second training stage was set to $128$ for efficient use of hardware.

\subsection{Main Results}
\label{sota}
In Table~\ref{tab:sota}, we evaluate $\texttt{QuantOD}$ with state-of-the-art methods for unsupervised outlier detection, all within the context of the inlier image classification task on CIFAR-10.~On average, we report superior performance, achieving a 15.84 and 19.91 percentage points improvement on FPR95 compared to the other leading methods, namely Energy~\cite{liu2020energy}, and Mahalanobis~\cite{lee2018simple}, respectively.~Our method also surpasses the AUROC score of Mahalanobis, with approximately three percentage points, while also delivering a better AUPR performance.~We achieve such improvements while maintaining the inlier classification accuracy when compared with other approaches.~With iSUN as the outlier dataset, improvements of 14.55 and 2.89 percentage points were recorded in FPR95 and AUROC, respectively, compared to the previous best method, i.e., ODIN~\cite{liang2018enhancing}.~Similarly, for Textures and SVHN as the outlier datasets, an improvement of 2.15 and 3.5 percentage points were observed in the FPR95 score, compared to the nearest previous approach, Mahalanobis.~We think the low-dimensional feature embeddings contain salient attributes of the inlier semantics.~Hence when trained on such inlier attributes, the flow estimates its probability distribution and can robustly distinguish an outlier feature from an inlier feature.~Our model inference runtime is comparable to the Energy method due to minimal computational overhead of the flow network.~We report faster runtime compared to the Mahalanobis, since this approach computes the Mahalanobis distance of the test features from the class-conditional Gaussians and selects the distance from the closest class as the confidence score.~Such an iterative computation result in a performance bottleneck.

\subsection{Ablation Studies}
\label{ablation}
We performed a detailed evaluation of our approach on varied experimental setups, such as the architecture of the classifier and normalizing flow network, the effect of mean vs. quantile-based $\mathcal{L}_{qNLL}$ loss,  quantifying outlier detection performance of $\texttt{QuantOD}$ with CIFAR-100 as inlier dataset and the performance comparison of $\texttt{QuantOD}$ with approaches that estimate class-conditional Gaussians.

\begin{figure}[!htb]
  \centering
  \includegraphics[width=0.97\linewidth]{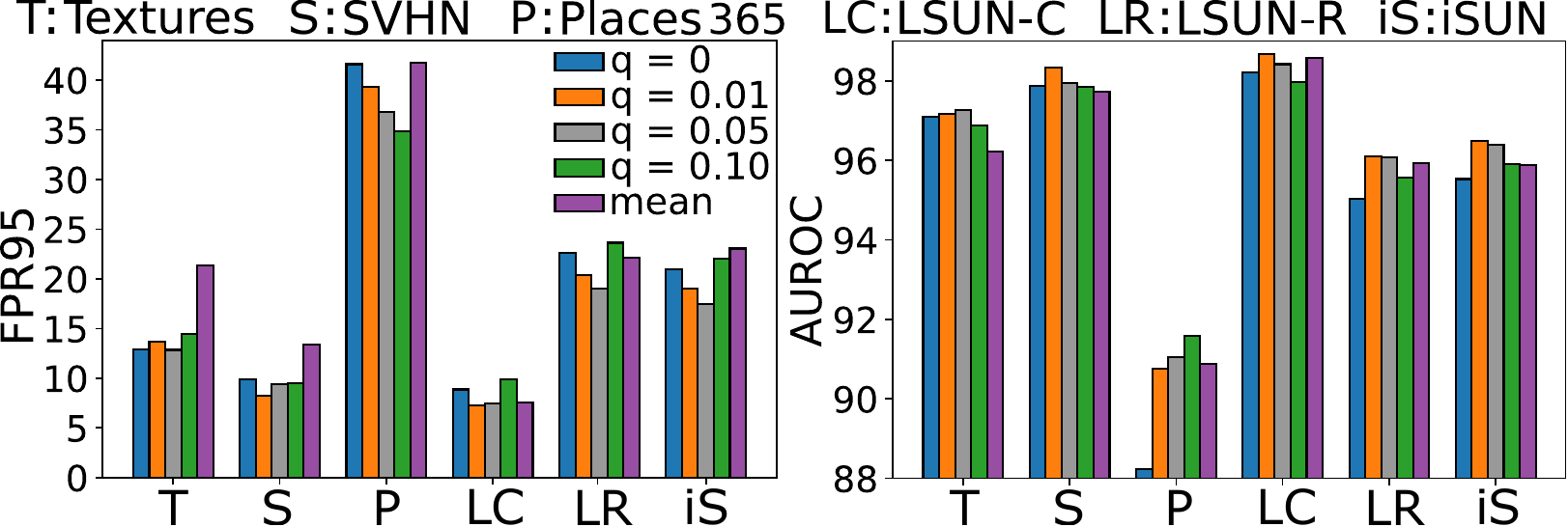}
   \caption[]{Validation on varying $q$ value in $\mathcal{L}_{qNLL}$ loss.}
   \label{vary q}
\end{figure}%
\begin{figure}[!htb]
  \centering
  \includegraphics[width=0.97\linewidth]{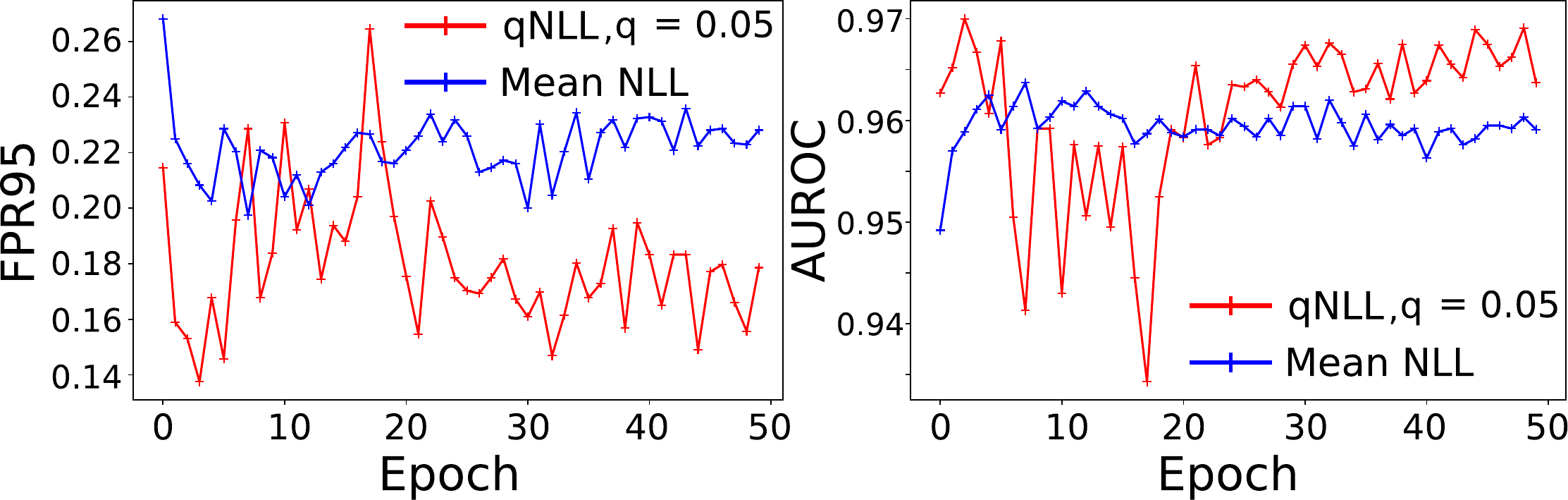}
   \caption[]{Validation of mean vs $q = 0.05$ at each training epoch.~The results are over iSUN as the outlier dataset.}
   \label{mean vs quantile}
\end{figure}

\paragraph{Mean vs Quantile-based $\mathcal{L}_{qNLL}$ Loss:}
We trained multiple instances of our $\texttt{QuantOD}$ framework on CIFAR-10 as an inlier by varying the parameter $q$ in the $\mathcal{L}_{qNLL}$ loss. Figure~\ref{vary q} presents the FPR95 and AUROC results on six outlier datasets. Generally, the model trained on the quantile-based negative log-likelihood loss with $q = 0.05$ outperforms the mean-based negative log-likelihood loss.~We believe the quantile-based loss 
encourages optimization of the likelihood scores
for inlier images situated at the boundary of the training manifold, 
leading to better outlier separation during inference.~Conversely, the mean-based loss may not maximize the likelihood of boundary inliers as effectively as those near the distribution mean.
To further examine this phenomenon, we compared the FPR95 and AUROC performance of mean vs 0.05-quantile based loss at each training epoch of the flow network. The results in Figure~\ref{mean vs quantile} demonstrate that as training progresses, the 0.05-quantile-based log-likelihood loss consistently enhances outlier detection performance.~In contrast, the mean-based loss reaches its peak after a few epochs and subsequently performs worse. This study validates the superiority of performing quantile-based maximum likelihood training of the generative model.   

\begin{table}[!htb]
    \centering
    \scalebox{0.9}{
    \begin{tabular}{|c|c|c|c|c|c|}
    \hline
        Classifier & Dim & Acc & FPR95  & AUROC & AUPR \\
        \hline  
         WideResNet40 & 128 & 94.68 & 9.39 & 97.94   & 99.59 \\
         ResNet18 & 512 & 95.10 & 31.01 &  94.85  & 98.99 \\
         \hline
    \end{tabular}}
    \caption{Validation of the classifier architecture trained on CIFAR-10 as the inlier and tested on SVHN.~The results are on Glow as the flow model.~The dimensionality of $r$ is denoted as Dim, and the classification accuracy as Acc in \%.}
    \label{classifier}
\end{table}

\paragraph{Architecture of the Classifier Backbone:}

We evaluated the classifier backbone for outlier detection performance by training them with CIFAR-10 as the inlier and testing on SVHN as the outlier dataset.~We kept the meta-parameters of the flow model fixed for this experiment.~We compared baseline model WideResNet40~\cite{DBLP:conf/bmvc/ZagoruykoK16} having the dimensionality of feature $r$ as 128 with ResNet18~\cite{7780459} having the feature dimensionality of 512.~Table~\ref{classifier} shows that the outlier detection performance reduces as the dimensionality of the features increase.~With higher dimensionality, the features become increasingly sparse, thereby preventing accurate modeling of inlier distribution and effective identification of outlier features.~The results convey that the low-dimensional features from the WideResNet40 model are more representative and therefore favored for outlier detection.~Additionally, increasing the dimensionality of $r$ does not significantly enhance classification accuracy, as the network architecture plays a more crucial role than the feature dimensions themselves.


\begin{table*}[!htb]
\centering
\scalebox{0.88}{
\begin{tabular}{|c|c|c|c|c|c|c|c|}
    \hline
    Flow Architecture & Blocks ($n$) & FC Layers & FC Neurons & Clamping & FPR95 $\downarrow$ (\%) & AUROC $\uparrow$ (\%) & AUPR $\uparrow$ (\%)  \\
    \hline
    NICE ~\shortcite{DBLP:journals/corr/DinhKB14} & 8 & 2 & 512 & - & 27.68 & 94.63 & 98.81  \\
    GIN ~\shortcite{Sorrenson2020Disentanglement} & 8 & 2 & 512 & 3.0 & 18.45 & 96.30 & 99.18  \\
    RealNVP ~\shortcite{dinh2017density} & 8 & 2 & 512 & 3.0 & 20.18 & 95.91 & 99.08  \\
    Glow ~\shortcite{kingma2018glow} & 8 & 2 & 512 & 3.0 & 17.17 & 96.19 & 99.15  \\
    \hline
    Glow ~\shortcite{kingma2018glow} & 2 & 2 & 512 & 3.0 & 19.26 & 95.89 & 99.09 \\
    Glow ~\shortcite{kingma2018glow} & 4 & 2 & 512 & 3.0 & 17.67 & 96.13 & 99.15  \\
    Glow ~\shortcite{kingma2018glow} & 16 & 2 & 512 & 3.0 & 17.28 & 96.36 & 99.21  \\
    \hline
    Glow ~\shortcite{kingma2018glow} & 8 & 1 & 512 & 3.0 & 19.80 & 95.69 & 99.02  \\
    Glow ~\shortcite{kingma2018glow} & 8 & 3 & 512 & 3.0 & 16.98 & 96.55 & 99.27  \\
    Glow ~\shortcite{kingma2018glow} & 8 & 4 & 512 & 3.0 & 17.03 & 96.39 & 99.23  \\
    \hline
    Glow ~\shortcite{kingma2018glow} & 8 & 2 & 128 & 3.0 & 17.83 & 96.35 & 98.20  \\
    Glow ~\shortcite{kingma2018glow} & 8 & 2 & 256 & 3.0 & 18.11 & 96.22 & 99.19  \\
    Glow ~\shortcite{kingma2018glow} & 8 & 2 & 1024 & 3.0 & 17.77 & 96.10 & 99.14 \\
    \hline
\end{tabular}}
\caption{Validation of the Normalizing Flow meta-parameters with WideResNet40 as the backbone and CIFAR-10 as the inlier dataset.~The results are averaged across the six outlier datasets.~The baseline flow network has 8 blocks, 2 layers, 512 neurons, and the Glow~\cite{kingma2018glow} architecture.~The parameter $q$ in the $\mathcal{L}_{qNLL}$ loss is fixed as 0.05 for all experiments.}\label{flow_net}
\end{table*}

\paragraph{Architecture of Normalizing Flow:}Table~\ref{flow_net} demonstrates the influence of altering the flow's meta-architecture.~We assess four different flow models, namely NICE~\cite{DBLP:journals/corr/DinhKB14}, RealNVP~\cite{dinh2017density}, Glow~\cite{kingma2018glow}, and GIN~\cite{Sorrenson2020Disentanglement}.~Notably, Glow achieves superior performance due to its precise density estimation.~Next, we examine the influence of varying the coupling blocks ($n$) in the Glow model.~The optimal results are observed with $n = 8$, as further increases offer negligible improvements in outlier detection performance but come with additional computational overhead.~For the evaluation of the flow after varying fully connected (FC) layers at each coupling block, Glow with eight coupling blocks is used.~The best performance is achieved with three FC layers, in terms of all the evaluation metrics.~We performed additional experiments by varying the number of neurons in the FC layers while keeping the number of layers and blocks fixed as two and eight, respectively.~Our results convey that 512 neurons in each of the FC layers provide the best performance for enhancing the flow model's outlier detection capability.
~By comparing the results in Table~\ref{tab:sota} and Table~\ref{flow_net}, it is evident that regardless of the meta-parameter and the architecture of the coupling block, our method outperforms most unsupervised outlier detection approaches.

\begin{table}[!htb]
    \centering
   \scalebox{0.87}{ \begin{tabular}{|c|c|c|c|c|c|}
    \hline
        Method & FPR95 & AUROC & AUPR & Acc & Time \\ \hline
        MSP~\shortcite{hendrycks2016baseline} & 80.41  & 75.53  & 93.93 & 75.96  & 12.81 \\ 
        ODIN~\shortcite{liang2018enhancing} & 74.64 & 77.43  & 94.23  & 75.96  & 17.17 \\
        Energy~\shortcite{liu2020energy}  & 73.60 & 79.56 & 94.87 & 75.96 & 12.69 \\ 
        QuantOD (Ours)  & 67.90 & 82.32 & 95.48 & 75.55 & 13.89 \\ \hline        
    \end{tabular}}
        \caption[]{Validation on CIFAR-100 as inlier and results averaged over six outlier datasets on WideResNet40 as the backbone.~The accuracy (Acc) in \% and Time in seconds.  } 
        \label{tab:cifar100}
\end{table}

\paragraph{CIFAR-100 as Inlier:}We further trained our $\texttt{QuantOD}$ framework on the CIFAR-100 as the inlier dataset and evaluated its performance with other unsupervised approaches on the six outlier datasets.~The results in Table~\ref{tab:cifar100} show that $\texttt{QuantOD}$ performs equally well when trained on CIFAR-100.~We report on average $5.7$ and $2.76$ percentage points improvement in FPR95 and AUROC performance when compared with the Energy-based method.~We also report marginal improvements in terms of AUPR metric and competitive inference runtime when compared with the Energy method.~The results convey the superiority of $\texttt{QuantOD}$ irrespective of the inlier dataset on which it is trained.

\begin{figure}[!htb]
  \centering
  \includegraphics[width=\linewidth]{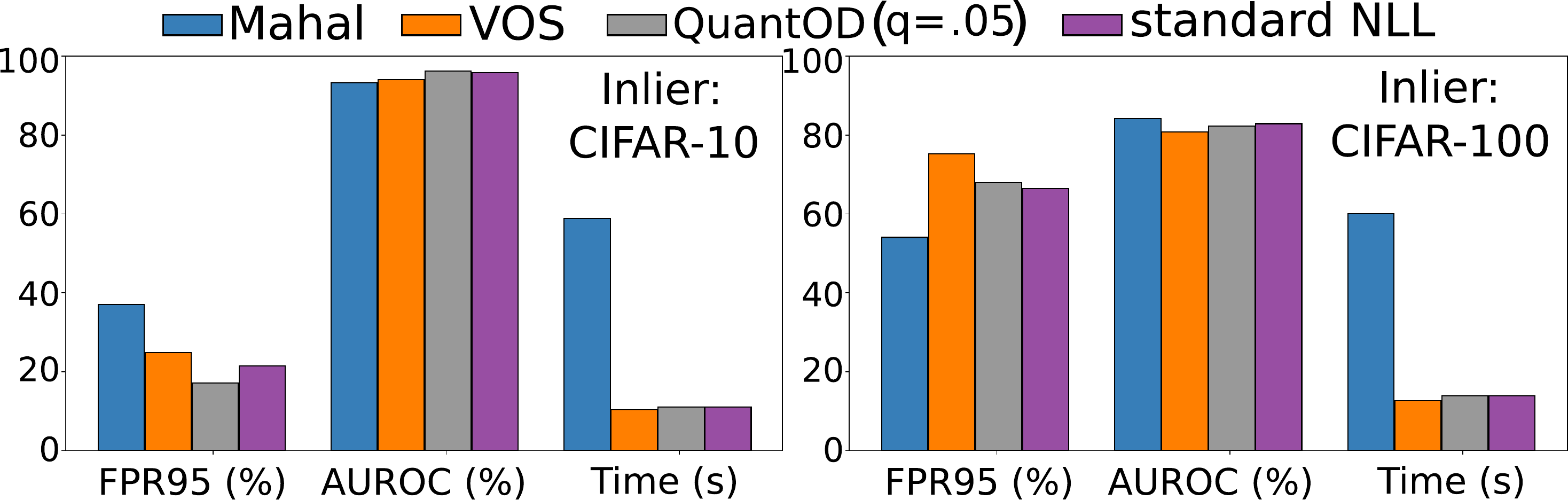}
   \caption[]{Comparison of \texttt{QuantOD} with approaches that perform class-conditional generative modeling of inlier features on both CIFAR-10 and CIFAR-100 as inliers.~The results are averaged over the six outlier datasets.}
   \label{quantod vs vos}
\end{figure}

\paragraph{Class-conditional Gaussians:}Unsupervised methods such as Mahalanobis~\cite{lee2018simple}, and self-supervised approaches like VOS~\cite{du2022vos}, rely on per-class generative modeling of inlier features.~Such a learning scheme is distinct from our approach, which employs flow model to learn a unified feature distribution across inlier classes.~Despite the notable difference, we compared the performance of $\texttt{QuantOD}$ with these two methods in~Figure~\ref{quantod vs vos}.~The results are based on the same WideResNet40 as the backbone architecture.~We report better outlier detection performance than both methods on inlier as CIFAR-10.~For the model trained on CIFAR-100, we report better FPR95 and AUROC scores than VOS while Mahalanobis performs better than our approach.~For a high number of inlier classes, Mahalanobis method has the better capability to estimate the closest distance of outlier features from class-conditional feature distribution.

\section{Conclusion}
Detecting shifts in data distribution is crucial to prevent neural networks from misclassifying unfamiliar inputs.~This work proposed a new maximum likelihood objective for robust outlier detection without needing outlier awareness during training.~The results demonstrate that, through the careful formulation of maximum likelihood training and accurate learning of inlier density in feature space, our approach outperforms other unsupervised methods and remains competitive with generative approaches that rely on per-class modelling.~Importantly, our framework adds a generative model to a discriminative classifier while incurring minimal computational overhead, making the approach suitable for real-time applications.~Future work can involve class-conditional normalizing flows, thereby aiming to improve the performance by leveraging full available supervision.

\appendix

\section{Acknowledgments}
This work primarily received funding from the German Helmholtz Association, the German Federal Ministry of Education and Research (BMBF) under Software  Campus (grant 01IS17044) and was supported by the Center for Scalable Data Analytics and Artificial Intelligence (ScaDS.AI) Dresden/Leipzig, Germany.~The work was also partially funded by DFG as part of TRR~248 -- CPEC (grant 389792660), Croatian Science Foundation (grant IP-2020-02-5851 ADEPT) and the Cluster of Excellence CeTI (EXC2050/1, grant 390696704).~The authors gratefully acknowledge the Center for Information Services and HPC (ZIH) at TU Dresden for providing computing resources.

\bibliography{aaai24}

\end{document}